\documentclass[letterpaper]{article} % DO NOT CHANGE THIS
\usepackage{aaai25}  % DO NOT CHANGE THIS
\usepackage{times}  % DO NOT CHANGE THIS
\usepackage{helvet}  % DO NOT CHANGE THIS
\usepackage{courier}  % DO NOT CHANGE THIS
\usepackage[hyphens]{url}  % DO NOT CHANGE THIS
\usepackage{graphicx} % DO NOT CHANGE THIS
\usepackage{natbib}  % DO NOT CHANGE THIS AND DO NOT ADD ANY OPTIONS TO IT
\usepackage{caption} % DO NOT CHANGE THIS AND DO NOT ADD ANY OPTIONS TO IT
\usepackage{algorithm}
\usepackage{algorithmic}

\usepackage{amsfonts,amssymb,amsmath}
\usepackage{xcolor}
\usepackage{multirow}
\usepackage{booktabs}

\usepackage{newfloat}
\usepackage{listings}
\DeclareCaptionStyle{ruled}{labelfont=normalfont,labelsep=colon,strut=off} % DO NOT CHANGE THIS
\floatstyle{ruled}
\newfloat{listing}{tb}{lst}{}
\floatname{listing}{Listing}
\title{TranSplat: Generalizable 3D Gaussian Splatting from \\ Sparse Multi-View Images with Transformers}
\author{
    Chuanrui Zhang\textsuperscript{\rm 1${\ast}$},  Yingshuang Zou\textsuperscript{\rm 1${\ast}$},  Zhuoling Li\textsuperscript{\rm 2},  Minmin Yi\textsuperscript{\rm 3},  Haoqian Wang\textsuperscript{\rm 1$\dag$}\\
}
\affiliations{
    \textsuperscript{\rm 1}Tsinghua University, ~\textsuperscript{\rm 2}The University of Hong Kong, ~\textsuperscript{\rm 3}E-surfing Vision Technology Co., Ltd \\
    \{zhang-cr22, zouys22\}@mails.tsinghua.edu.cn, ~ lizhuoling@connect.hku.hk, \\
    yimm18@chinatelecom.cn, ~ wanghaoqian@tsinghua.edu.cn
}

\begin{document}

% \maketitle

\twocolumn[{%
    \renewcommand\twocolumn[1][]{#1}%
    \maketitle
    % \vspace{-1.5cm}
    \includegraphics[width=0.95\textwidth]{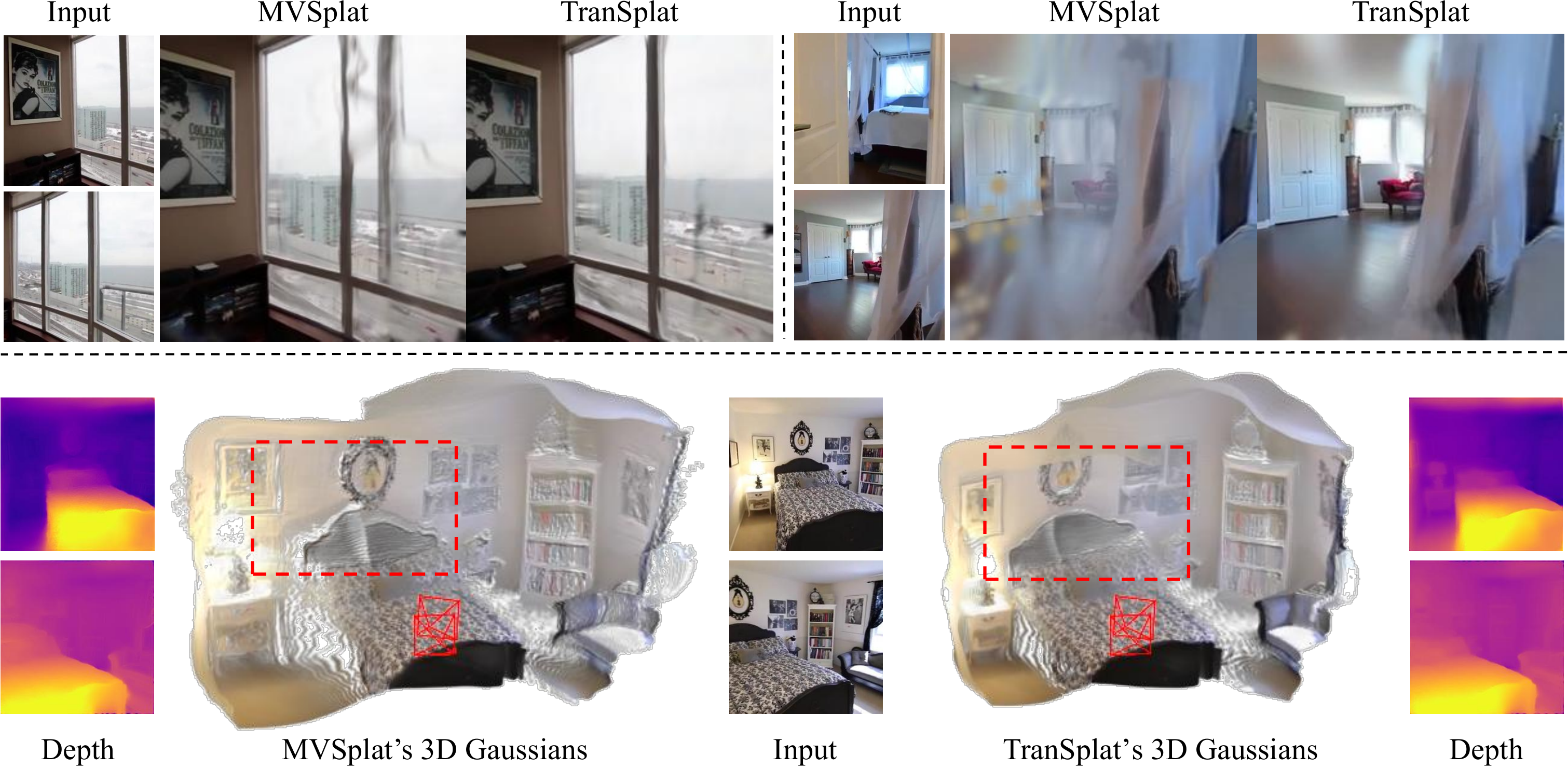}
    % \vspace{-6pt}
    \captionof{figure}
    {
    Given two-view images, TranSplat achieves higher quality in both novel view synthesis and 3D Gaussian construction, particularly in challenging areas such as low-texture, repetitive patterns, and non-overlapping regions, compared to MVSplat~\cite{chen2024mvsplat}. Our novel view synthesis results contain fewer artifacts and maintain better geometric consistency.
    }
    \label{fig:teaser}
    \vspace{12pt}
}]
{\let\thefootnote\relax\footnotetext{{$^{\ast}$ Equal contribution.  $\dag$ Corresponding authors.}}}

\begin{abstract}
Compared with previous 3D reconstruction methods like Nerf, recent Generalizable 3D Gaussian Splatting (G-3DGS) methods demonstrate impressive efficiency even in the sparse-view setting. However, the promising reconstruction performance of existing G-3DGS methods relies heavily on accurate multi-view feature matching, which is quite challenging. Especially for the scenes that have many non-overlapping areas between various views and contain numerous similar regions, the matching performance of existing methods is poor and the reconstruction precision is limited. To address this problem, we develop a strategy that utilizes a predicted depth confidence map to guide accurate local feature matching. In addition, we propose to utilize the knowledge of existing monocular depth estimation models as prior to boost the depth estimation precision in non-overlapping areas between views. Combining the proposed strategies, we present a novel G-3DGS method named TranSplat, which obtains the best performance on both the RealEstate10K and ACID benchmarks while maintaining competitive speed and presenting strong cross-dataset generalization ability. Our code, and demos will be available at: https://xingyoujun.github.io/transplat.

\end{abstract}

\vspace{-0.4cm}
\section{Introduction}

% 1. 可泛化的稀疏视点重建任务目的是从稀疏的2D图像中恢复真实3D结构，playing a fundamental role in various applications.
% 2. 之前很多的工作关注于使用NeRF来表示3D结构。受限于 ray-marching-based volume rendering的固有限制，使得基于NeRF的工作无法做到实时渲染。最近得益于3DGS极快的渲染速度以及高质量的渲染结果，目前受到了越来越多的关注。3DGS固有地避免了NeRF昂贵的体积采样过程，从而实现了高效、高质量的3D重建和新颖的视图合成。
% The task of generalizable sparse-view reconstruction, which aims to recover the 3D structure from a sparse set of 2D images without per-scene optimization, is crucial in many applications. 
% Remarkable works has been done using several different neural scene representations.
% Many previous works~\cite{yu2021pixelnerf} focus on employing Neural Radiance Fields (NeRF)~\cite{mildenhall2021nerf} to represent 3D structures. 
% However, the inherent limitations of the ray-marching-based volume rendering method hinder the efficient real-time rendering capabilities of NeRF-based approaches. 
% Recently, the 3D Gaussian Splatting (3DGS) ~\cite{kerbl3Dgaussians} technique has garnered increasing attention due to its exceptionally rapid rendering speed and the high quality of its rendering performances.
% Unlike NeRF, which relies on computationally expensive ray-marching, 3DGS represents the scene with a set of 3D Gaussians, which are initialized with point clouds from Structure from Motion (SfM).
% This approach is more effectient bypasses the dense volumetric ray-marching process, leading to efficient and high-quality 3D reconstruction as well as the synthesis of novel views.

The task of generalizable sparse-view 3D reconstruction, which aims to recover 3D structure from a sparse set of images without scene-specific optimization, is crucial in many applications like virtual reality~\cite{li2023voxelformer, zou2024m2}. Significant progress has been achieved using various neural scene representations, such as NeRF \cite{mildenhall2021nerf, wang2022attention} and 3D Gaussian Splatting (3DGS) \cite{kerbl3Dgaussians, lu2024scaffold, liu2023animatable}. Unlike NeRF, which relies on computationally expensive ray-marching \cite{xu2022pointnerf, gao2024general}, 3DGS represents the scene with a set of 3D Gaussians benefiting from rasterization-based rendering. These characteristics make 3DGS more efficient because it bypasses the dense volumetric ray-marching process and enables both efficient and high-quality 3D reconstruction and novel view synthesis.

However, traditional 3D reconstruction and novel view synthesis methods mostly require scene-specific optimization \cite{barron2021mip, pons2021dnerf}, resulting in increased computational demands and longer processing times. Recently, there has been growing interest in Generalizable 3D Gaussian Splatting (G-3DGS) for sparse-view reconstruction. Some methods \cite{charatan2024pixelsplat, chen2024mvsplat} have emerged that are capable of reconstructing 3D scenes in a single forward pass without requiring extra optimization. These advancements achieve high-quality reconstruction results while reducing computational overhead, thereby enhancing the efficiency and applicability of 3D reconstruction across various contexts. For example, PixelSplat \cite{charatan2024pixelsplat} and MVSplat \cite{chen2024mvsplat} unproject depth maps from corresponding views to serve as the centers of 3D Gaussians. However, the performances of these methods rely on accurate pixel-level matching based on depth, which is challenging to realize and thus limits the reconstruction precision of these methods. We observe that especially in the scenes scenes with occlusions, insufficient texture, or repetitive patterns, the matching result is quite unsatisfactory~\cite{yao2018mvsnet, sun2021loftr, pautrat2023gluestick, wang2022mvster}. Additionally, the reconstruction of non-overlapping areas between views is also challenging for existing methods due to the lack of matching pairs.

% 在这篇文中，我们提出了一种新的稀疏视角可泛化3D重建框架，names Transplat，去预测pixel-align的gaussian parameters。the key insight of Transplat 是我们认为通过扩大匹配feature的感受野可以有效的提升feature 的匹配能力。
% In this paper, we propose a novel framework for 3D reconstruction, named Transplat, which predicts pixel-aligned Gaussian parameters and enables efficient rendering for novel view synthesis. 
% The key insight of Transplat is that we believe expanding the receptive field of features when matching can significantly enhance their matching capabilities and inject the strong neural prior from pre-trained depth features can help improve the network's generalization ability.
To address the aforementioned limitations, we propose \textbf{TranSplat}, a novel framework for generalizable sparse-view 3D reconstruction. 
TranSplat recovers 3D structures from sparse views by projecting each pixel into 3D Gaussian primitives leveraging the predicted depths and features.
Therefore, accurate depth prediction is crucial to guarantee geometric consistency across 3D Gaussians from multiple perspectives.
% Since our approach avoids scene-specific optimization, maintaining geometric consistency across multi-view 3D Gaussians is crucial, making accurate depth prediction essential.
To precisely estimate 3D Gaussian centers, we generate depth estimates using a transformer-based module called the Depth-aware Deformable Matching Transformer (DDMT). This module prioritizes depth candidates with high confidence based on an initial depth distribution, which is calculated by the Coarse Matching module through assessing cross-view feature similarities. 
Additionally, TranSplat utilizes monocular depth priors and employs the Depth Refine U-Net to further refine the depth distribution. 
With the refined depth and image features, our method predicts all Gaussian parameters—center, covariance, opacity, and spherical harmonics coefficients—in parallel for each pixel.
% To precisely estimate 3D Gaussian centers, we divide the depth estimation process into two phases: coarse-to-fine matching and depth refinement. 
% In the coarse-to-fine matching phase, we first calculate cross-view feature similarities for all depth candidates to obtain a coarse depth distribution. We then refine this distribution using a transformer-based module, Depth-Aware Deformable Matching (DDM), which prioritizes depth candidates with high confidence based on the initial distribution.
% In the depth refinement phase, TranSplat leverages monocular depth priors and employs the Depth Refine U-Net to further refine the depth distribution. With the refined depth and image features, our method predicts all Gaussian parameters (center, covariance, opacity, and spherical harmonics coefficients) in parallel for each pixel. 
TranSplat is trained end-to-end using only ground-truth images for supervision.

We evaluate TranSplat on two large-scale benchmarks: RealEstate10K~\cite{zhou2018re10k} and ACID~\cite{liu202acid}. Extensive experiments are conducted and demonstrate that TranSplat achieves the best results in G-3DGS. Notably, compared to existing counterparts~\cite{charatan2024pixelsplat, chen2024mvsplat}, TranSplat presents strong cross-dataset generalization ability.

% contributions
Comprehensively, our main contributions are as follows:
\begin{itemize}
\vspace{-0.05cm}
\item We propose to utilize the depth confidence map to enhance matching between various views and correspondingly significantly improve the reconstruction precision in regions with insufficient texture or repetitive patterns.
\vspace{-0.05cm}
\item We propose a strategy that encodes the priors of monocular depth estimators into the prediction of Gaussian parameters, ensuring precise 3D Gaussian centers are estimated even in non-overlapping areas.
\vspace{-0.05cm}
\item The derived method TranSplat achieves the best results on two large-scale benchmarks and presents strong cross-dataset generalization ability.

\end{itemize}
\begin{figure*}[htp]
\centering
\includegraphics[width=0.95\textwidth]{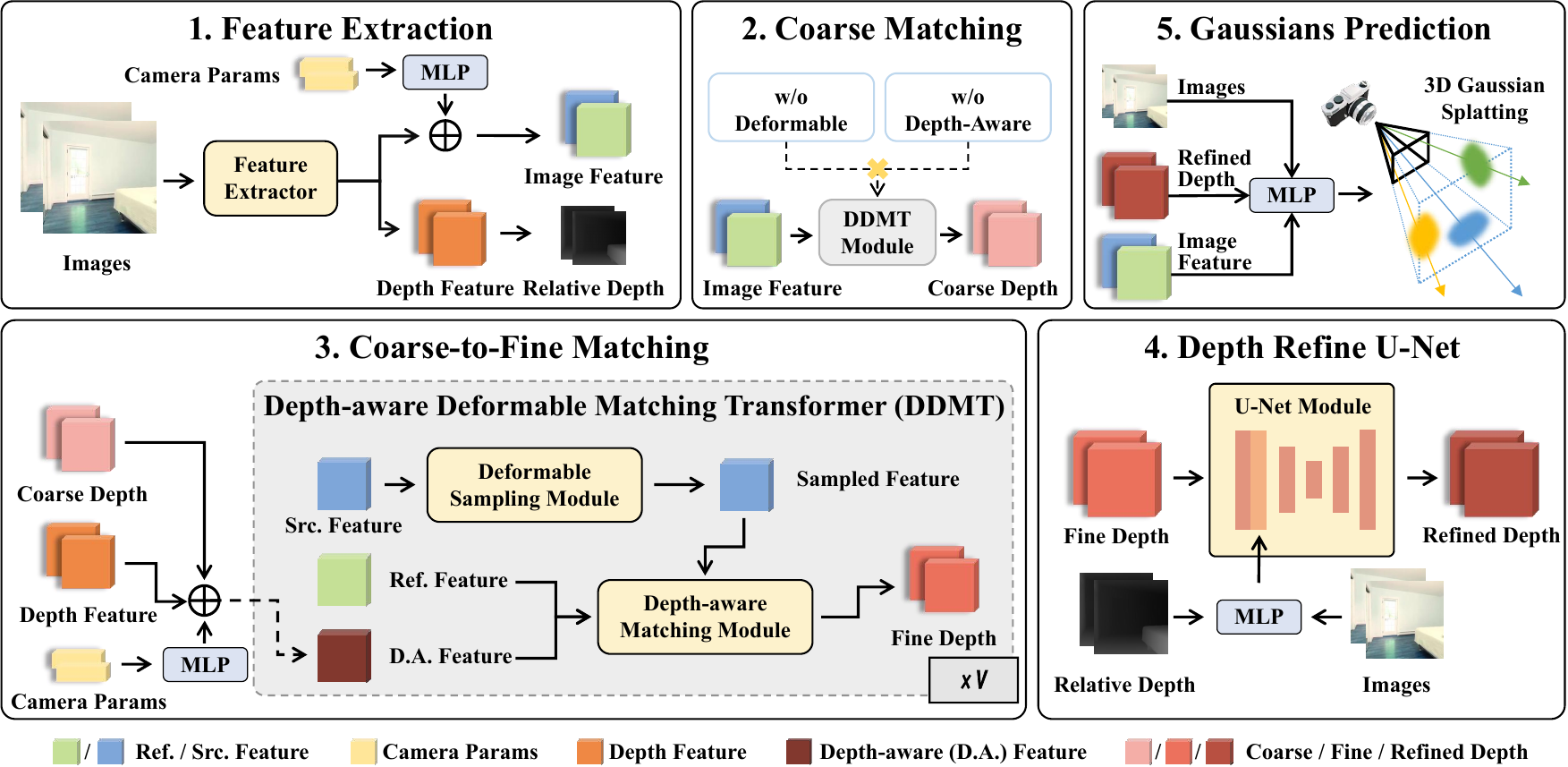} % Reduce the figure size so that it is slightly narrower than the column.
\caption{\textbf{Framework of TranSplat.} 
Our method takes multi-view images as input and first extracts image features and monocular depth priors. Next, the coarse-to-fine matching stage is used to obtain a geometry-consistent depth distribution for each view. Specifically, we compute multi-view feature similarities using our proposed Depth-Aware Deformable Matching Transformer module. 
The Depth Refine U-Net is then employed to further refine the depth prediction. Finally, we predict pixel-wise 3D Gaussian parameters to render novel views.
}
\label{fig:pipeline}
\vspace{-0.5cm}
\end{figure*}

\section{Related Work}

% \vspace{-0.1cm}
\subsection{Sparse-view Scene Reconstruction}
With advancements in 3D scene reconstruction technologies such as NeRF and 3D Gaussian Splatting (3DGS), there is growing interest in reconstructing scenes from sparse inputs~\cite{fan2024instantsplat, guo2024depth, ni2024colnerf, yu2021pixelnerf}. The limited overlap in sparse inputs poses a significant challenge to scene reconstruction \cite{shi2024zerorf, irshad2023neo}.
Existing sparse-view methods are primarily categorized into two main approaches: per-scene optimization methods and feed-forward inference methods. The former typically leverages multi-view geometry constraints to jointly optimize rendering results and camera poses~\cite{niemeyer2022regnerf, truong2023sparf, deng2022depth_supervised_NeRF}. These methods may also improve novel-view results by incorporating additional prior knowledge, such as depths~\cite{chung2024depth_reg_gaussian, li2024dngaussian}. However, per-scene optimization methods require iterative processes to achieve the final 3D representation.
In contrast, feed-forward inference methods reconstruct the entire scene in a single feed-forward pass, eliminating the need for subsequent optimization. These methods benefit from powerful priors from large datasets, which enables them to generalize effectively across different datasets.

\vspace{-0.1cm}
\subsection{3DGS for Generalizable Reconstruction}
Currently, 3DGS-based generalizable reconstruction methods are attracting significant interest due to their rapid rendering speed and superior cross-dataset generalization capabilities. For example, PixelSplat~\cite{charatan2024pixelsplat} employs a multi-view epipolar transformer to accurately infer the real scale of scenes, effectively resolving the inherent scale ambiguity in real-world datasets. It predicts Gaussian parameters through dense probability distributions, which facilitates accurate and efficient 3D structure estimation.
MVSplat~\cite{chen2024mvsplat} reconstructs 3D scenes by constructing a cost volume that enables joint prediction of depth and Gaussian parameters. The centers of the Gaussians are determined by unprojecting the depth map into 3D space.
LatentSplat~\cite{wewer2024latentsplat} addresses reconstruction challenges in object-centric scenes with 360° views by introducing variational Gaussians. It uses a lightweight VAE-GAN decoder to generate RGB images of novel views.
Despite these advancements, 3DGS-based methods still face challenges, such as achieving robust reconstructions in areas with occlusions, lack of texture, or repetitive patterns.

\vspace{-0.1cm}
\subsection{Multi-view Stereo}
Multi-View Stereo (MVS) techniques enable the reconstruction of 3D structures from 2D images~\cite{collins1996space_sweep}. Recent learning-based MVS methods~\cite{yao2018mvsnet, gu2020casmvsnet, ding2022transmvsnet, peng2022unimvsnet, xu2023unifying} utilize 3D cost volumes to predict depth for each view and then fuse these depth maps to obtain the final point cloud. These methods typically require dense depth maps for effective model training.
In contrast, our proposed method directly extracts the 3D representation from sparse views, eliminating the need for separate depth estimation and point cloud fusion processes. Notably, our approach requires only RGB images for supervision, simplifying data acquisition and reducing reliance on costly depth sensors.
% \vspace{-0.2cm}
\section{Method}
\label{sec:method}
We introduce \textbf{TranSplat}, a novel generalizable sparse-view scene reconstruction network that employs a transformer-based architecture. The overview of our proposed method is illustrated in Figure \ref{fig:pipeline}. Given $ K $ sparse-view images $ \{\mathbf{I}^i\}_{i=1}^K $ (where $ \mathbf{I}^i \in \mathbb{R}^{H \times W \times 3} $) and their corresponding camera parameters, our objective is to predict a 3D Gaussian representation of the scene through a single forward pass. The Gaussian parameters include position $ \mu $, opacity $ \alpha $, covariance $ \Sigma $, and color $ c $ (represented as spherical harmonics). Using these parameters, we can render any novel views.

% \textit{We commence by discussing the feature extractor in Section \ref{sec:method_fe}. Subsequently, we employ coarse matching to obtain the initial coarse depth in Section \ref{sec:method_cm}. Next, we introduce our proposed Depth-Aware Deformable Matching (DDM) module and the depth refinement U-Net module in Sections \ref{sec:method_fm} and \ref{sec:method_dr}, respectively. In Section \ref{sec:method_gp}, we elucidate the Gaussian parameter prediction module. Finally, we provide the training loss functions in Section \ref{sec:method_tl}.}

\vspace{-0.2cm}
\subsection{Feature Extraction}
\label{sec:method_fe}
We employ a standard CNN and Transformer architecture \cite{xu2022gmflow,xu2023unifying} to extract multi-view image features $ \{\mathbf{F}^i\}_{i=1}^K $ (where $ \mathbf{F}^i \in \mathbb{R}^{\frac{H}{4} \times \frac{W}{4} \times C} $), with $ C $ denoting the channel dimension. Notably, we incorporate camera parameters into each CNN feature using a squeeze-excitation (SE) layer \cite{hu2018squeeze} to provide global spatial information. In our self- and cross-attention layers, we adopt the local window attention mechanism of the Swin Transformer \cite{liu2021swin} to enhance information exchange between different views. Furthermore, we utilize the large vision model DepthAnythingV2 \cite{depth_anything_v2} to obtain monocular depth priors. We extract the last layer of the DepthAnythingV2 module as depth features $ \{\mathbf{D}^i\}_{i=1}^K $ (where $ \mathbf{D}^i \in \mathbb{R}^{\frac{H}{4} \times \frac{W}{4} \times C} $), as well as the relative depth outputs $ \{\mathbf{d}^i_{rel.}\}_{i=1}^K $ (where $ \mathbf{d}^i_{rel.} \in \mathbb{R}^{H \times W} $).

% \vspace{-0.05cm}
\subsection{Coarse Matching}
\label{sec:method_cm}
Following feature extraction, acquiring an initial depth distribution is essential for generating depth confidence maps. To simplify the code implementation, the DDMT module is utilized directly, bypassing the need for deformable sampling and depth-aware matching. This approach is taken because those models rely on depth information to produce attention maps for depth candidates.
% Following feature extraction, the image features $ \{\mathbf{F}^i\}_{i=1}^K $ are passed through the lightweight version of our proposed Depth-Aware Deformable Matching (DDM) module. This lightweight version of the DDM module, which omits deformable sampling and depth-aware matching, facilitates a rapid match to obtain the coarse depth as the initial depth distribution for the subsequent stage.

Consider the source view $ i \in K $ and the target view $ j \in K $ (where $ i \neq j $) as an example. We first construct a depth candidate $\mathbf{d}_{cand.}^i \in \mathbb{R}^{\frac{H}{4} \times \frac{W}{4} \times D}$ using the plane-sweep stereo approach \cite{yao2018mvsnet}. Then, we sample the $j$-th view feature $\mathbf{F}^{ij}_{sample} \in \mathbb{R}^{\frac{H}{4} \times \frac{W}{4} \times D \times C}$ as follows:
\vspace{-0.1cm}
\begin{equation}
\mathbf{F}^{ij}_{sample} = \textrm{Sample}(\mathbf{F}^j, \mathbf{P}^i, \mathbf{P}^j, \mathbf{d}_{cand.}^i),
\end{equation}
% \vspace{-0.1cm}
where $\textrm{Sample}$ denotes the sampling operation \cite{xu2023unifying}, and $\mathbf{P}^i$ and $\mathbf{P}^j \in \mathbb{R}^{4 \times 4}$ represent the camera projection matrices from the $i$-th and $j$-th views to the world coordinate system. $D$ denotes the depth dimension.

Subsequently, we compute the dot product between $\mathbf{F}^i$ and $\mathbf{F}^{ij}_{sample}$ to obtain the coarse depth $\mathbf{d}^i_{coarse} \in \mathbb{R}^{\frac{H}{4} \times \frac{W}{4} \times D}$:

\vspace{-0.6cm}
\begin{equation}
\mathbf{d}^i_{coarse} = \frac{\mathbf{F}^i \otimes \mathbf{F}^{ij}_{sample}}{\sqrt{C}},
\end{equation}
where $\otimes$ represents the element-wise product, and the results are summed along the channel dimension. When more than two views are provided as input, we apply the Coarse Matching module to the other views and compute the pixel-wise average across all views to obtain the final coarse depth.

\vspace{-0.1cm}
\begin{figure}[!t]
\centering
\includegraphics[width=\linewidth]{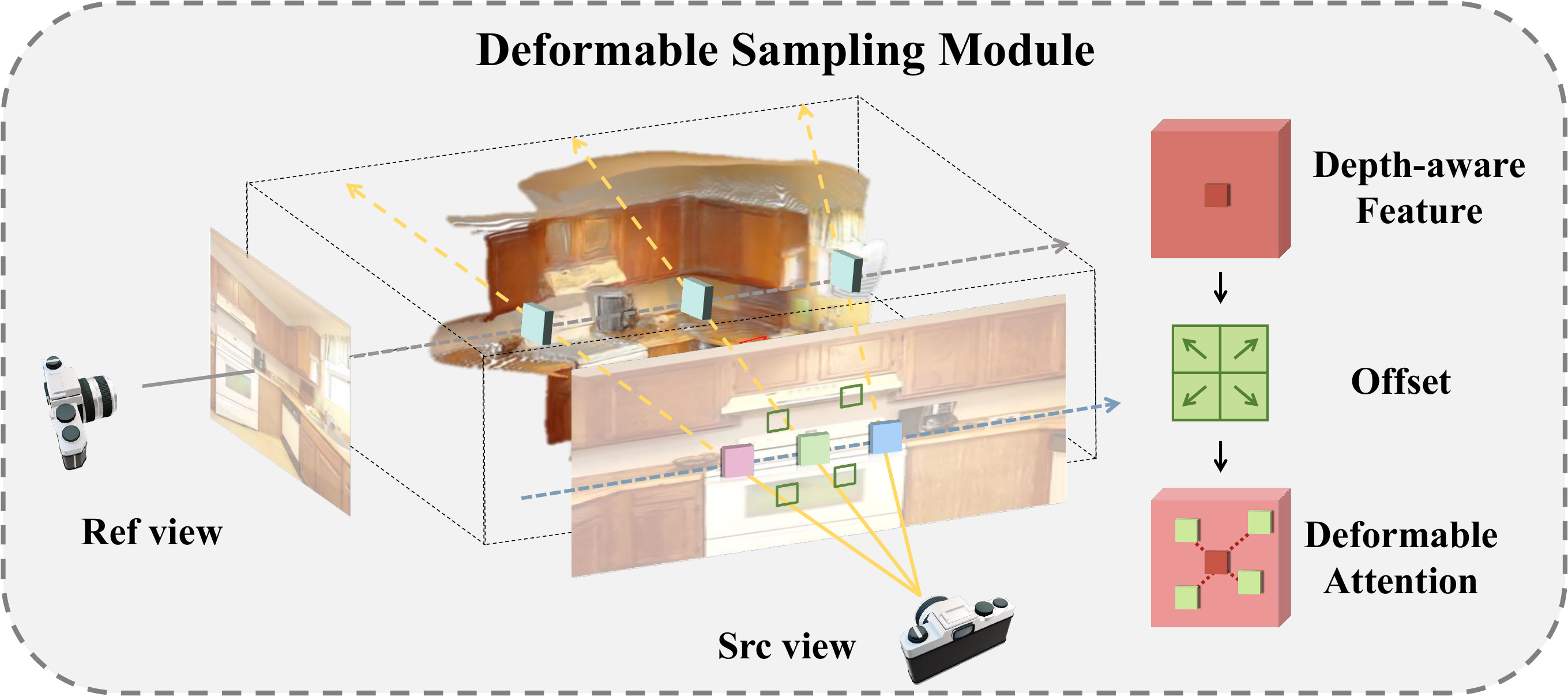}
\caption{\textbf{The details of Deformable Sampling module.} 
We sample cross-view points using deformable attention, which enhances the aggregation of local spatial information.
}
\label{fig:DWM}
\vspace{-0.5cm}
\end{figure}

\vspace{-0.1cm}
\subsection{Coarse-to-Fine Matching}
\label{sec:method_fm}
\begin{figure}[!t]
\centering
\includegraphics[width=\linewidth]{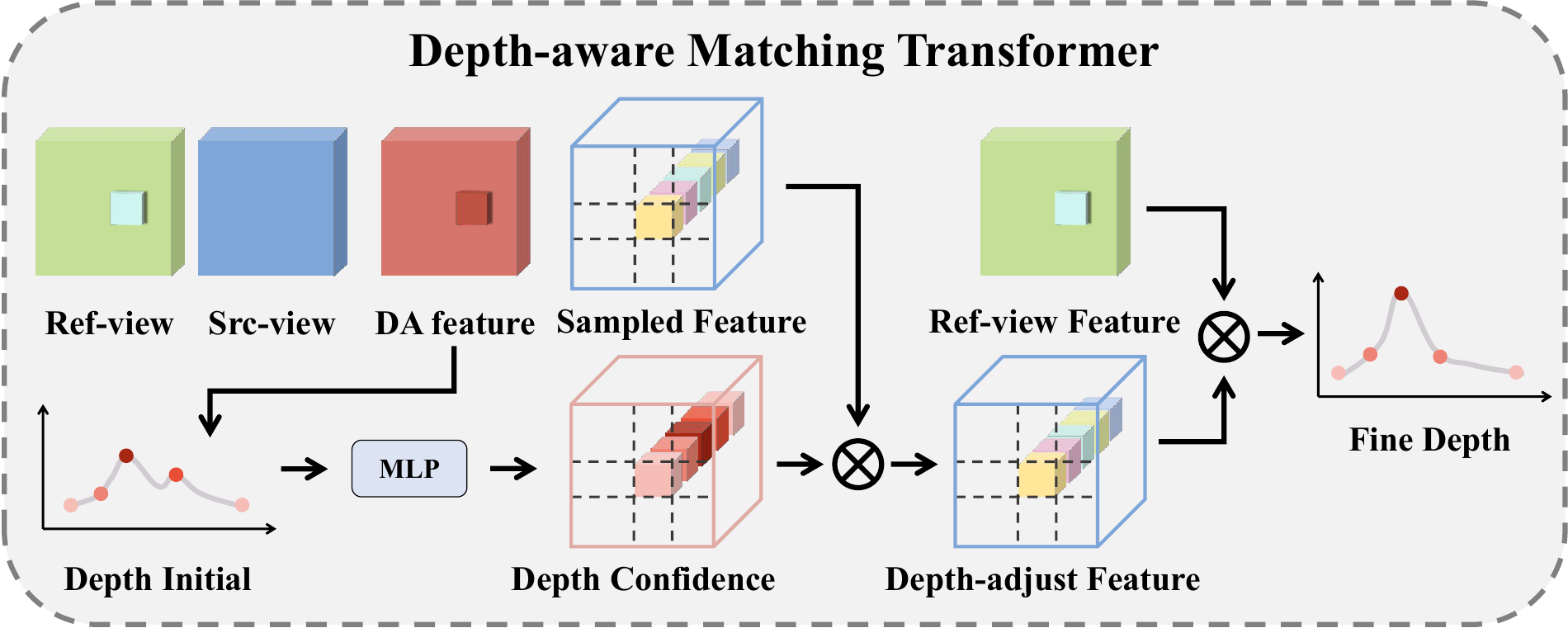}
\caption{\textbf{The illustration of Depth-aware Matching Transformer.}
We utilize the depth confidence map to integrate into our Depth-aware Matching Transformer, enabling the network to prioritize regions with higher depth confidence and improving the overall depth prediction accuracy. (Here D.A. feature represents depth-aware feature.)
}
\label{fig:DAM}
\vspace{-0.5cm}
\end{figure}

Since the Coarse Matching module only performs pixel-aligned feature matching, it struggles with low-texture areas and repetitive patterns. To address this issue, we designed the Deformable Sampling (DS) module (in Figure \ref{fig:DWM}), which aggregates the spatial information of local features. Furthermore, we observed that the Coarse Matching module equally prioritizes all depth candidates. To overcome this limitation, we propose the Depth-aware Matching Transformer (DMT) module (in Figure \ref{fig:DAM}) to guide our network to focus on the correct depth candidates.

Similar to Section \ref{sec:method_cm}, we first consider the source view $ i $ and the target view $ j $. To construct the depth-aware feature (d.a. feature) $\mathbf{F}^i_{d.a} \in \mathbb{R}^{\frac{H}{4} \times \frac{W}{4} \times C}$, we fuse the coarse depth map $\mathbf{d}^i_{coarse}$, depth feature $\mathbf{D}^i$, and camera parameters $\mathbf{P}^i$ through the depth-aware network:

\vspace{-0.2cm}
\begin{equation} 
\mathbf{F}^i_{d.a} = f_{\theta}(\mathbf{d}^i_{coarse}) + f_{\phi}(\mathbf{D}^i, \mathbf{P}^i),
\end{equation}
% \vspace{-0.1cm}
where $f_{\theta}$ represents a Multi-Layer Perceptron (MLP) layer, and $f_{\phi}$ denotes an SE layer employed to inject position information into the depth feature.

Next, we apply the Deformable Sampling module to obtain the deformable sampled feature $\mathbf{F}^{ij}_{d.s.} \in \mathbb{R}^{\frac{H}{4} \times \frac{W}{4} \times D \times C}$ of the $j$-th view:

\vspace{-0.2cm}
\begin{equation} 
\mathbf{F}^{ij}_{d.s.} = \textrm{D.S.}(\mathbf{F}^j, \mathbf{P}^i, \mathbf{P}^j, \mathbf{d}^i_{cand.}, \Delta \mathbf{p}^i),
\end{equation}
% \vspace{-0.1cm}
where $\textrm{D.S.}$ represents the Deformable Sampling module, and $\Delta \mathbf{p}^i \in \mathbb{R}^{\frac{H}{4} \times \frac{W}{4} \times D \times P}$ represents the image-level sampling offsets, with $P$ as the number of deformable points.

Our proposed Depth-Aware Matching module contains self- and cross-attention layers to enhance depth prediction. The coarse depth $\mathbf{d}^i_{coarse}$ is first input into the self-attention layer to aggregate local depth information. Next, we calculate the depth residual $\Delta \mathbf{d}^i \in \mathbb{R}^{\frac{H}{4} \times \frac{W}{4} \times D}$ in the cross-attention layer:

\vspace{-0.25cm}
\begin{equation} 
\Delta \mathbf{d}^i = \frac{\mathbf{F}^i \otimes \sum_P{(\mathbf{W}^i \otimes \mathbf{F}^{ij}_{d.s.})}}{\sqrt{C}},
\end{equation}
% \vspace{-0.1cm}
where $\otimes$ represents the element-wise product, and the results are summed along the channel dimension. $\mathbf{W}^i \in \mathbb{R}^{\frac{H}{4} \times \frac{W}{4} \times D \times P}$ represents the depth-aware attention weights. The sampling offsets and attention weights are predicted through the offset layer $f_p$ and the attention layer $f_w$, respectively:

\vspace{-0.2cm}
\begin{equation} 
\Delta \mathbf{p}^i = f_{p}(\mathbf{F}^i_{d.a.}), \quad \mathbf{W}^i = f_{w}(\mathbf{F}^i_{d.a.}).
\end{equation}

The fine depth is then computed as follows:

\vspace{-0.1cm}
\begin{equation} 
\mathbf{d}^i_{fine} = \mathbf{d}^i_{coarse} + \Delta \mathbf{d}^i.
\end{equation}

Similar to Section \ref{sec:method_cm}, we get the fine depth of $K$ views separately when there are more than two views as input.

\vspace{-0.1cm}
\subsection{Depth Refine U-Net}
\label{sec:method_dr}
Recent methods~\cite{charatan2024pixelsplat, chen2024mvsplat} primarily focus on cross-image information to infer depth. However, as the number of input views decreases, there are large areas without cross-view matches, resulting in unreliable depth outputs for those regions. To optimize the depth distribution of these non-overlapping areas, we propose providing strong monocular depth priors.

We designed the Depth Refine U-Net module to combine the accurate geometric consistency from the matching results with the reliable relative depth from monocular depth priors. Inspired by diffusion models~\cite{rombach2022high, duan2023diffusiondepth}, our Depth Refine U-Net module receives input images and relative depths as conditions to refine the input depths. 

Taking input images $ \{\mathbf{I}^i\}_{i=1}^K $, relative depths $ \{\mathbf{d}^i_{rel.}\}_{i=1}^K $, and fine depths $ \{\mathbf{d}^i_{fine}\}_{i=1}^K $ as input, our module outputs per-view depth residuals:
\vspace{-0.2cm}
\begin{equation} 
\mathbf{d}^i_{refine} = \mathbf{d}^i_{fine} + f_{r}(\mathbf{d}^i_{fine}, \mathbf{I}^i, \mathbf{d}^i_{rel.}),
\end{equation}
where $f_{r}$ denotes our Depth Refine U-Net module and $\{\mathbf{d}^i_{refine}\}_{i=1}^K (\mathbf{d}^{i}_{refine} \in \mathbb{R}^{H \times W})$ represents the final refined depths.

\vspace{-0.2cm}
\subsection{Gaussian Prediction}
\label{sec:method_gp}
In this section, we leverage our depth-aware features to predict the parameters of a set of 3D Gaussian primitives $\{(\mu_i, \alpha_i, \Sigma_i, c_i)\}_{i=1}^M$ that parameterize the scene. We set $M = H \times W \times K$ following \cite{charatan2024pixelsplat} as we predict a Gaussian primitive per pixel from $K$ views.

\noindent \textbf{Gaussian center $\mu$.} 
We utilize the final refined depth $\mathbf{d}^i_{refine}$ predictions to directly unproject every pixel to global 3D points using the camera parameters. These points are then directly selected as the centers of the 3D Gaussians.

\noindent \textbf{Opacity $\alpha$.} 
For each pixel, the opacity should be set to 1 if the depth prediction is accurate, as the points must lie on the surface. The opacity can be regarded as the confidence of the depth prediction. Therefore, we can predict opacity using a simple MLP layer with the refined depth prediction as input.

\noindent \textbf{Covariance $\Sigma$ and Color $c$.} 
Similar to other 3D Gaussian approaches~\cite{kerbl3Dgaussians, chen2024mvsplat}, we calculate color from the predicted spherical harmonic coefficients and use a scaling matrix $s$ and a rotation matrix $R(\theta)$ to represent the covariance matrix $\Sigma$.

\vspace{-0.2cm}
\begin{equation} 
\Sigma = R(\theta)^T \text{diag}(s) R(\theta).
\end{equation}

These parameters are predicted using MLP layers.

\subsection{Training Loss}
\label{sec:method_tl}
Equipped with the predicted 3D Gaussian parameters, we can render images from novel views. We directly supervise the quality of the novel RGB images using Mean Squared Error (MSE) and Learned Perceptual Image Patch Similarity (LPIPS) losses~\cite{zhang2018unreasonable} as in \cite{chen2024mvsplat}.

\begin{figure*}[t]
\centering
\includegraphics[width=0.9\textwidth]{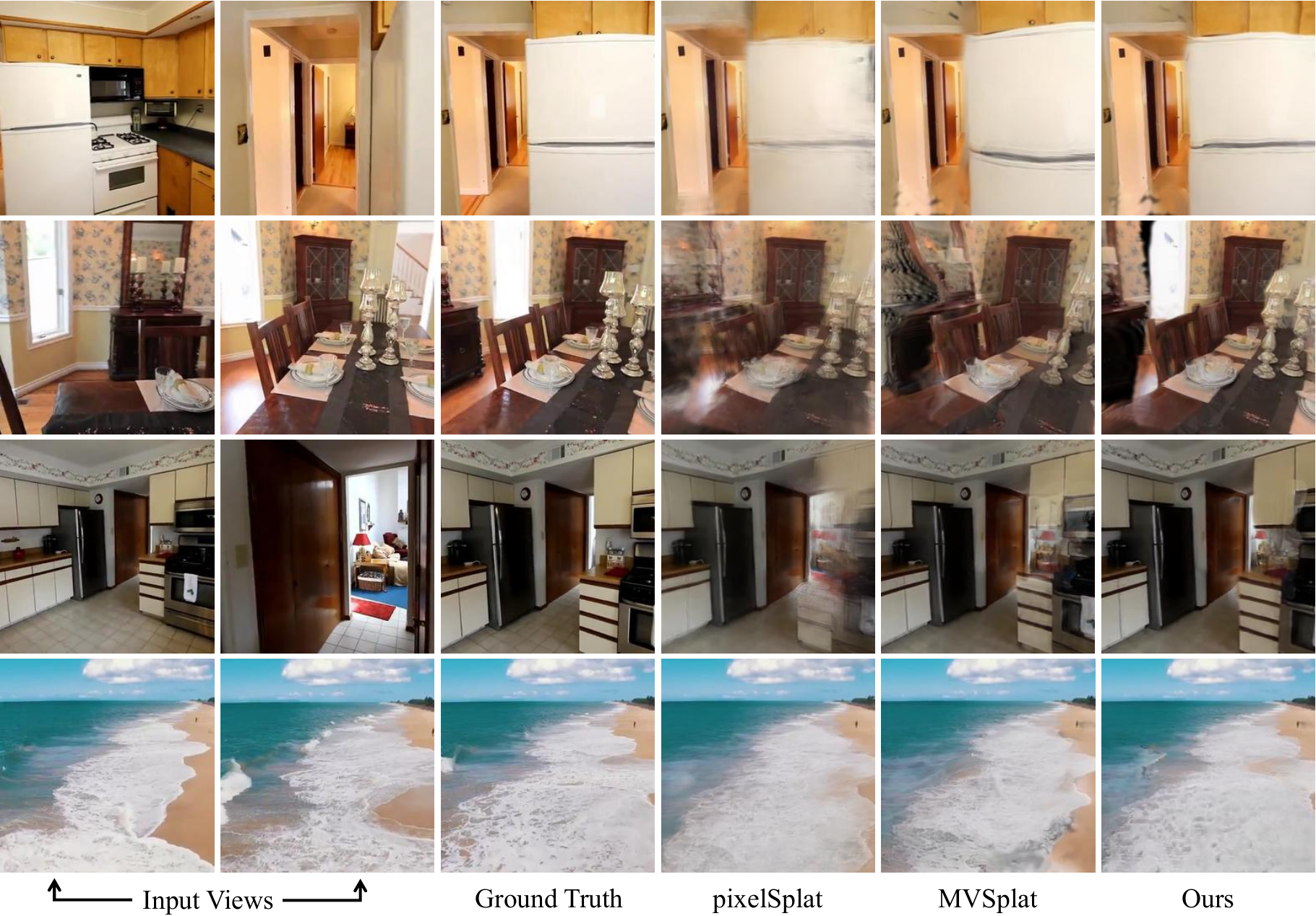} % Reduce the figure size so that it is slightly narrower than the column.
\caption{
\textbf{The qualitative comparisons with SOTA methods.}
Our method outperforms other state-of-the-art methods across various scenes, with the first three rows from RealEstate10K and the last one from ACID. TranSplat excels in challenging regions thanks to the effectiveness of our transformer-based Depth-Aware Deformable Matching Transformer (DDMT).
}
\label{fig:sota_comparsion}
\end{figure*}

\begin{table*}[!t]
\centering
\resizebox{0.9\linewidth}{!}{
\begin{tabular}{l|ccc|ccc|cc}
\toprule
\multirow{2}{*}{\textbf{Method} }&\multicolumn{3}{c|}{RealEstate10K} &\multicolumn{3}{c|}{ACID}      &\multicolumn{2}{c}{Consumption}  \\
     & PSNR$\uparrow$ & SSIM$\uparrow$ & LPIPS$\downarrow$ & PSNR$\uparrow$ & SSIM$\uparrow$ & LPIPS$\downarrow$ & Memory$\downarrow$ & Time$\downarrow$     \\
\midrule
pixelSplat(~\citeauthor{charatan2024pixelsplat})  & 25.89   &0.858  &0.142  & 28.14 & 0.839 & 0.150 & 5319 MB & 0.1553 s  \\
MVSplat(~\citeauthor{chen2024mvsplat}) & 26.39   & 0.869  & \underline{0.128}  & \underline{28.25} & \underline{0.843} & \underline{0.144} & \underline{2906 MB} & \textbf{0.0667 s} \\
Ours 200k   & \underline{26.44}  & \underline{0.871}  & \underline{0.128}  & 28.20     & \underline{0.843} & 0.145   & \textbf{2228 MB}     & \underline{0.0851 s} \\
Ours 300k   & \textbf{26.69}  & \textbf{0.875}  & \textbf{0.125}  & \textbf{28.35}     & \textbf{0.845}    & \textbf{0.143}      & \textbf{2228 MB}     & 0.0872 s \\
\bottomrule
\end{tabular}
}
\caption{\textbf{Quantitative comparisons SOTA methods.}
Our TranSplat outperforms others on large-scale benchmarks, offering competitive inference speed and memory efficiency. Remarkably, it achieves the best performance with only 200K iterations (66\% of 300K) in the RealEstate10K dataset, a testament to the effectiveness of our transformer-based architecture. (\textbf{Bold} figures indicate the best and \underline{underlined} figures indicate the second best) }
\label{tab:all_results}
\vspace{-0.2cm}
\end{table*}

\begin{figure}[t]
\centering
\includegraphics[width=\linewidth]{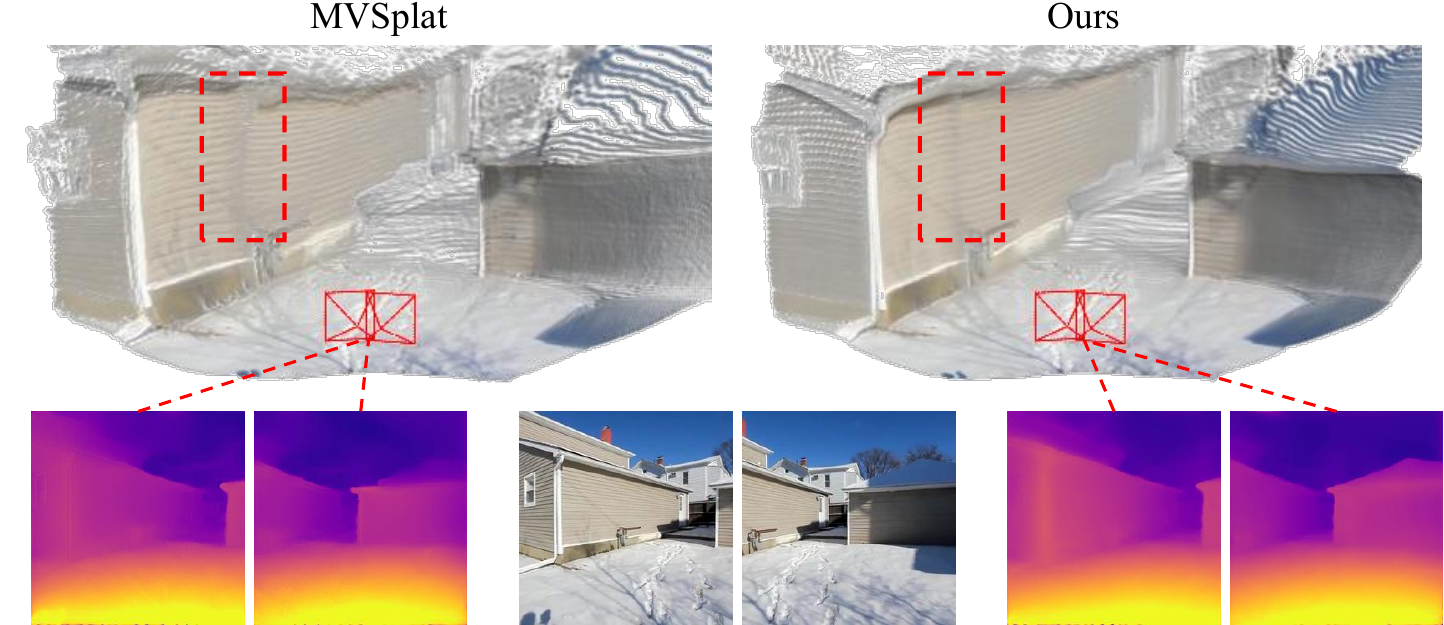}
\caption{The qualitative comparisons on 3D Gaussians.
TranSplat generates significantly higher-quality 3D Gaussians in scenes with repetitive patterns (e.g., walls) and non-overlapping areas (e.g., the right roof), showcasing the effectiveness of our proposed modules.
}
\label{fig:pc_result}
\vspace{-0.1cm}
\end{figure}

\begin{table}[!h]
% \vspace{0.1cm}
\centering
\resizebox{\linewidth}{!}{
\begin{tabular}{l|ccc|ccc}
\toprule
\multirow{2}{*}{\textbf{Method}}& \multicolumn{3}{c|}{Re10k$\rightarrow$DTU} & \multicolumn{3}{c}{Re10k$\rightarrow$ACID}\\
& PSNR$\uparrow$ & SSIM$\uparrow$ & LPIPS$\downarrow$ & PSNR$\uparrow$ & SSIM$\uparrow$ & LPIPS$\downarrow$  \\
\midrule
pixelSplat  & 12.89    & 0.382   & 0.560   & 27.64 & 0.830  & 0.160      \\
MVSplat & 13.94    & 0.473   & 0.385   & \underline{28.15}  & \underline{0.841} & \underline{0.147} \\
Ours 100k   & \underline{14.72}   &  \underline{0.517}   & \underline{0.344}   & {27.73}      & {0.834}      & {0.155}     \\
Ours 300k   & \textbf{14.93}   &  \textbf{0.531}   & \textbf{0.326}   & \textbf{28.17}      & \textbf{0.842}      & \textbf{0.146}      \\
\bottomrule
\end{tabular}
}
\caption{\textbf{Cross-dataset generalization results.}
We conduct zero-shot tests on the DTU dataset~\cite{jensen2014dtu} using models trained on RealEstate10K~\cite{zhou2018re10k} and ACID~\cite{liu202acid} without any fine-tuning. TranSplat demonstrates substantial generalization ability, outperforming other state-of-the-art methods on the DTU dataset, even with only 100K iterations of training.
}
\vspace{-0.1cm}
\label{tab:cross_dataset}
\end{table}

\begin{figure}[!h]
\centering
\includegraphics[width=\linewidth]{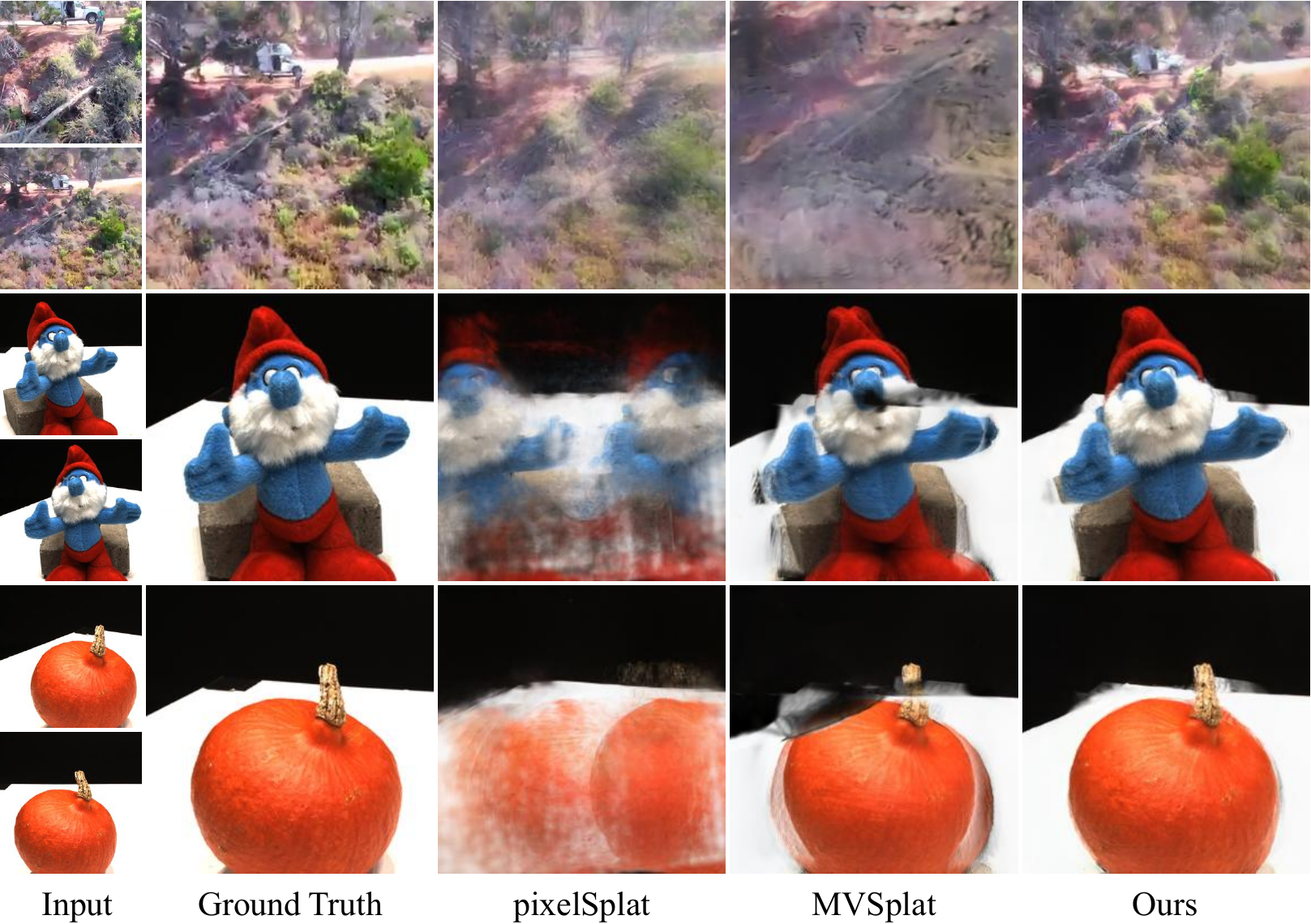}
\caption{\textbf{The qualitative comparisons of cross-dataset generalization.}
Using a model trained on RealEstate10K, we directly render scenes from the ACID dataset (first row) and the DTU dataset (last three rows). TranSplat produces images with notable quality improvements, demonstrating its ability to adapt to diverse depth distributions and effectively handle low-texture regions without requiring additional training.
}
\label{fig:cross_dataset}
\vspace{-0.3cm}
\end{figure}

\vspace{-0.2cm}
\section{Experiments}

\subsection{Experimental Setup}
\noindent \textbf{Datasets.}
We train and evaluate our model using the large-scale RealEstate10K~\cite{zhou2018re10k} and ACID~\cite{liu202acid} datasets. RealEstate10K comprises home walkthrough videos from YouTube, with 67,477 scenes for training and 7,289 scenes for testing. The ACID dataset, featuring aerial landscape videos, includes 11,075 training scenes and 1,972 testing scenes. Both datasets provide camera poses computed by SFM software. In line with MVSplat~\cite{chen2024mvsplat}, we train our model using two context views and evaluate all methods from three novel target viewpoints for each test scene. Additionally, to assess cross-dataset generalization, we evaluate all methods on the multi-view DTU dataset~\cite{jensen2014dtu}, selecting 16 validation scenes with four novel views each.

% \noindent \textbf{Metrics.}
% To obtain quantitative results, we compare each method's novel RGB images with ground-truth frames by computing the Peak Signal-to-Noise Ratio (PSNR), Structural Similarity Index (SSIM)~\cite{wang2004image}, and Perceptual Distance (LPIPS)~\cite{zhang2018unreasonable}.

\noindent \textbf{Metrics.}
To obtain quantitative results, we compare the novel RGB images produced by each method with ground-truth frames using the Peak Signal-to-Noise Ratio (PSNR), Structural Similarity Index (SSIM)\cite{wang2004image}, and Perceptual Distance (LPIPS)\cite{zhang2018unreasonable}.

\noindent \textbf{Implementation details.}
Input images are resized to 256 × 256, following the method outlined in~\cite{chen2024mvsplat}. In all experiments, the number of depth candidates is set to 128. We sample $P=4$ deformable points in the Depth-Aware Deformable Matching Transformer for the main results. For the DepthAnythingV2~\cite{depth_anything_v2} module, we use the base size to balance training cost and result quality. All models are trained with a batch size of 14 on 7 RTX 3090 GPUs for 300,000 iterations using the Adam~\cite{kingma2014adam} optimizer. During inference, we measure speed and memory cost with one RTX 3090 GPU.

\vspace{-0.1cm}
\subsection{Main Results}
\noindent\textbf{Novel view synthesis quality.}
We report quantitative results in Table \ref{tab:all_results}. Our evaluation compares TranSplat with state-of-the-art (SOTA) methods on scene-level novel view synthesis from sparse views. TranSplat outperforms the baselines across all metrics, with competitive inference speed (0.06s vs. 0.08s). 
% PixelSplat~\cite{charatan2024pixelsplat} utilizes an epipolar-transformer-based approach, while MVSplat~\cite{chen2024mvsplat} employs a cost-volume-based method. 
Under identical training conditions, TranSplat not only exceeds all previous SOTA methods in visual quality, as indicated by the PSNR and SSIM metrics but also shows improvements in the LPIPS metric, which better aligns with human perception. Notably, TranSplat achieves superior performance compared to other SOTA methods with only 200K iterations (66\% of 300K), demonstrating that our transformer-based architecture enhances convergence efficiency by focusing on accurate depth candidates.

The qualitative comparisons of the three methods are visualized in Figure \ref{fig:sota_comparsion}.
TranSplat achieves superior quality on novel view images across various scenes. The first row showcases a scene with a low-texture area, such as the "fridge surface," where our method excels. This improvement is due to our Depth-Aware Deformable Matching Transformer, which aids the network in aggregating long-term spatial information and concentrating on the correct areas through attention mechanisms. Our method also demonstrates superiority in mismatched scenes, such as the "in-room area" in the second row and the "lamp" in the third row. This enhancement is attributable to our Depth Refine U-Net, which utilizes monocular depth priors to refine Gaussian predictions in non-overlapping areas. Similar results are evident in the ACID datasets, as shown in the fourth row.

\noindent\textbf{Reconstruction quality.}
TranSplat generates impressive 3D Gaussian primitives compared to the recent state-of-the-art method MVSplat~\cite{chen2024mvsplat}, as shown in Figure \ref{fig:pc_result}. Our TranSplat method can produce high-quality depth maps in challenging scenes, such as those with repetitive patterns (e.g., "the walls"). Notably, our method significantly outperforms in generating 3D Gaussians in non-overlapping areas (e.g., "the roof on the right"). These results collectively demonstrate the superiority of our proposed method.

\vspace{-0.1cm}
\subsection{Cross-dataset generation}
Our proposed TranSplat demonstrates significant superiority in generalizing to \textit{out-of-distribution} novel scenes, as illustrated in Table \ref{tab:cross_dataset}. 
Following the settings of MVSplat~\cite{chen2024mvsplat}, we directly use the model trained on RealEstate10K (an indoor dataset) to test on ACID (an outdoor dataset) and DTU (an object-centric dataset). The results on the ACID dataset are nearly identical, as ACID consists of aerial landscape scenes that may be reduced to an image interpolation task. TranSplat outperforms other state-of-the-art methods on the DTU dataset by a substantial margin across all metrics. Even with only 100K (33\% of 300K) iterations of training, our method exhibits excellent generalization in cross-dataset settings. This superiority is attributed to our Depth-Aware Deformable Matching Transformer, which efficiently adjusts the attention area as the dataset's scale changes. Additionally, the monocular depth prior contributes to more generalized Gaussian parameter predictions across different multi-view camera distributions. We present additional results in Figure \ref{fig:cross_dataset}. TranSplat inherently improves rendering quality in novel views, even when scenes differ significantly in appearance. In contrast, the results of MVSplat contain artifacts, indicating poor geometric consistency. Our TranSplat method provides reliable depth information for mismatched areas, whereas MVSplat tends to force-fit the depth distribution of the training dataset.

\begin{table}[htb]
\centering
\resizebox{0.8\linewidth}{!}{
\begin{tabular}{ccccccc}
\toprule
 C.M.  & DDMT & D.R.U. & C.P.E. & PSNR$\uparrow$ & SSIM$\uparrow$ & LPIPS$\downarrow$  \\
\midrule
         &  $\checkmark$    &    &     & 25.011        & 0.841         & 0.152             \\
 $\checkmark$       &     &    &     & 25.020        & 0.842         & 0.150             \\
 $\checkmark$       & $\checkmark$   &    &     & 25.231              & 0.846             & 0.148                  \\
 $\checkmark$       & $\checkmark$   & $\checkmark$  &     & 25.516         & 0.853              & 0.145                  \\
 $\checkmark$       & $\checkmark$   & $\checkmark$  & $\checkmark$   & \textbf{25.615}        & \textbf{0.855}        & \textbf{0.142}              \\
\bottomrule
\end{tabular}}
\caption{
Ablation study on the RealEstate10K~\cite{zhou2018re10k} dataset. C.M. and DDM refer to Coarse Matching and the Depth-Aware Deformable Matching module, respectively. D.R.U. and C.P.E. stand for Depth Refine U-Net and Camera Parameter Encoding. The ablation results demonstrate the effectiveness of our proposed methods, with the final line representing our complete model.
}
\vspace{-0.3cm}
\label{tab:ablation_tab}
\end{table}

\vspace{-0.1cm}
\subsection{Ablation Study}
To evaluate the effect of our different modules, we conduct ablation experiments on the RealEstate10K~\cite{zhou2018re10k} dataset with a batch size of 9, trained for 100K iterations. All experiments show that our proposed modules contribute to performance improvements. Additional hyper-parameter ablation details are provided in the Supplements.

\vspace{0.1cm}
\noindent\textbf{Importance of Depth-aware Deformable Matching Transformer.}
The Depth-Aware Deformable Matching Transformer (DDMT) is responsible for maintaining multi-view geometric consistency. The DDMT module requires an initial depth estimate to generate depth-adjusted features for efficient matching. We present the results in Table \ref{tab:ablation_tab}. To assess the importance of the DDMT and the influence of different initial depths, we conduct ablation experiments with random depth estimates (1st row), only coarse matching (2nd row), and coarse-to-fine strategies (3rd row). We find that using random depth estimates as the initial input causes the DDMT module to fail in generating a meaningful depth confidence map, resulting in performance similar to that of simply using the coarse matching module. However, when providing coarse depth estimates, the DDMT module improves results, as the model can focus on depth-informative areas.

\noindent\textbf{Importance of Depth Refine U-Net.}
We use the Depth Refine module to optimize the depth map by incorporating monocular depth information from the DepthAnythingV2~\cite{depth_anything_v2}. To investigate the necessity of this module, we perform an ablation study without the Depth Refine U-Net (4th row in Table \ref{tab:ablation_tab}). We observe performance improvements with the Depth Refine module, as it provides reliable depth information for non-overlapping areas.

\noindent\textbf{Importance of Camera Parameter Encoding.}
We investigate the significance of Camera Parameter Encoding (C.P.E.) for feature extraction by performing an ablation study on its use (5th row in Table \ref{tab:ablation_tab}). The results demonstrate that C.P.E. is essential for injecting global spatial information into feature maps, leading to improved performance.

\noindent\textbf{Ablation of multi-view number.}
As discussed in Section \ref{sec:method}, our proposed TranSplat is designed to process multi-view images. Our method benefits from additional input views, as more matchable areas are provided. We test our multi-view capability directly on the DTU dataset using a model trained on the RealEstate10K dataset with 2-view inputs. Compared to the 2-view results presented in Table \ref{tab:cross_dataset}, TranSplat achieves a PSNR of 15.14, an SSIM of 0.561, and an LPIPS of 0.330 with 3-view inputs, and a PSNR of 15.22, an SSIM of 0.570, and an LPIPS of 0.340 with 4-view inputs. We observe that TranSplat's performance improves as the number of input views increases, even without additional training.

\section{Conclusion}
In this work, we introduce TranSplat, a novel generalizable sparse-view scene reconstruction network with a transformer-based architecture. Equipped with multi-view inputs, our method predicts a set of 3D Gaussian primitives to represent the scene. To address incorrect matches caused by similar areas, we propose the Depth-Aware Deformable Matching Transformer, which enhances attention to accurate depth distribution. To improve 3D Gaussians in non-overlapping areas, we designed the Depth Refine module, which optimizes the depth map using monocular depth priors. Our method achieves state-of-the-art (SOTA) performance on two large-scale scene-level reconstruction benchmarks and demonstrates superior cross-dataset generalization.
\noindent\textbf{Limitations:} We predict 3D Gaussian primitives per pixel, which can result in fewer points on the sides of objects, potentially causing gaps when rotating around objects. It would be interesting to explore methods to generate more meaningful 3D Gaussians for the sides of objects.

% \clearpage

\bibliography{aaai25}

\clearpage
\section{Appendix-Ablation Study}
% \vspace{0.5cm}
% \subsubsection{Importance of Deformable Sampling}
\subsection{Importance of Deformable Sampling}
Table \ref{tab:depth_sample} presents a comparison between methods with and without Deformable Sampling (D.S.). The experimental results consistently demonstrate that employing the deformable sampling strategy leads to improved outcomes.

\begin{table}[ht]
\centering
\resizebox{0.7\linewidth}{!}{
\begin{tabular}{c|ccc}
\toprule
Method &PSNR$\uparrow$ & SSIM$\uparrow$ & LPIPS$\downarrow$  \\
\midrule
Base   & \textbf{25.615} & \textbf{0.855} & \textbf{0.141} \\
w/o D.S. & \underline{25.527} & \underline{0.852} & \underline{0.144} \\
\bottomrule
\end{tabular}
}
\caption{\textbf{Ablation of Deformable Sampling.} Experimental results demonstrate that our design effectively enhancing the quality of novel view synthesis. (\textbf{Bold} figures indicate the best and \underline{underlined} figures indicate the second best)}
\label{tab:depth_sample}
\end{table}

% \subsubsection{Ablation of the size of monocular depth module}
\subsection{Ablation of the Depth Prior Module Size}
Table \ref{tab:depth_anything} shows that our model's performance declines when the size of the DepthAnythingV2~\cite{depth_anything_v2} module is altered. We hypothesize that the model requires a balance between geometric consistency from matching and depth priors from monocular models. Consequently, any change in the depth prior, whether an improvement or degradation, hinders the model's ability to converge as effectively as before.

\begin{table}[h]
\centering
\resizebox{0.7\linewidth}{!}{
\begin{tabular}{c|ccc}
\toprule
Method & PSNR$\uparrow$ & SSIM$\uparrow$ & LPIPS$\downarrow$ \\
\midrule
Base     & \textbf{25.615} & \textbf{0.855} & \textbf{0.141} \\
D.A. S   & 25.313 & 0.848 & \underline{0.146} \\
D.A. L   & \underline{25.426} & \underline{0.850} & \underline{0.146} \\
\bottomrule
\end{tabular}
}
\caption{\textbf{Ablation of the DepthAnythingV2 Module Size.} We replace the base monocular depth priors with small version (D.A. S) and large version (D.A. L) of the DepthAnythingV2 model ~\cite{depth_anything_v2}. The experimental results demonstrate that the DepthAnythingV2 base model delivers the best performance. (\textbf{Bold} figures indicate the best and \underline{underlined} figures indicate the second best)
}
\label{tab:depth_anything}
\end{table}

% \vspace{2cm}

% \subsubsection{Ablation of the number of DDMT module}
\subsection{Ablation of the number of DDMT module}
Table \ref{tab:number_match} illustrates that increasing the number of Depth-Aware Deformable Matching Transformer (DDMT) modules does not improve the quality of novel view synthesis. We hypothesize that repeating the DDMT module diminishes performance by causing the model to overfocus on local areas.

\begin{table}[h]
\centering
\resizebox{0.7\linewidth}{!}{
\begin{tabular}{c|ccc}
\toprule
DDMT & PSNR$\uparrow$ & SSIM$\uparrow$ & LPIPS$\downarrow$ \\
\midrule
0   & 25.264 & 0.846 & \underline{0.149} \\
1   & \textbf{25.343} & \textbf{0.848} & \textbf{0.148}\\
2   & \underline{25.285} & \underline{0.847} & \underline{0.149} \\
3   & 25.282 & 0.846 & \underline{0.149} \\
\bottomrule
\end{tabular}
}
\caption{\textbf{Ablation of the number of DDMT module.} We conduct ablation studies on the number of DDMT modules. The experimental results indicate that employing the DDMT module once produces the best results. (\textbf{Bold} figures indicate the best and \underline{underlined} figures indicate the second best)
}
\label{tab:number_match}
\end{table}

\section{Appendix-Visualized}

\subsection{More Results in RealEstate10K}
\begin{figure*}[htp]
\centering
\includegraphics[width=0.95\textwidth]{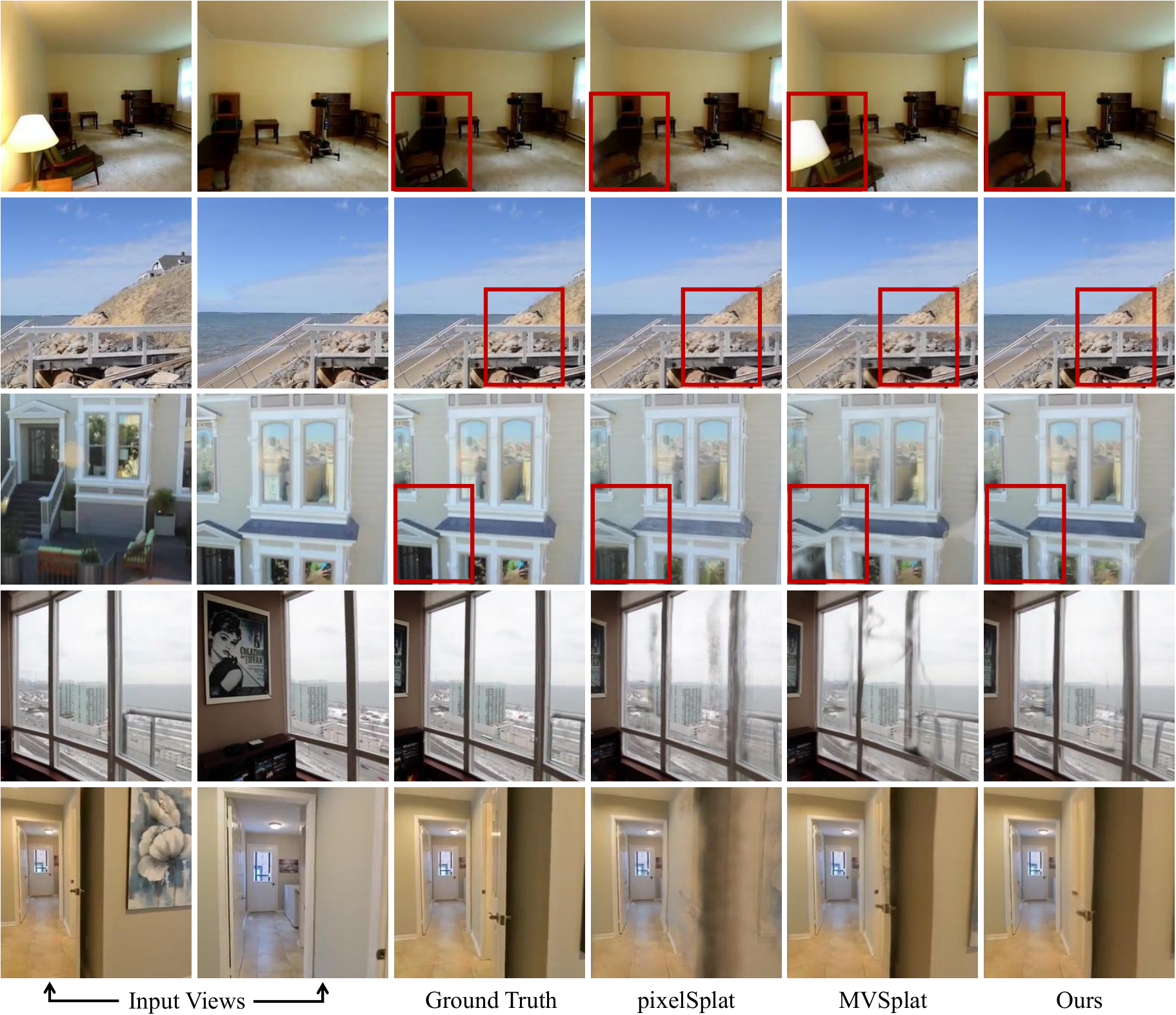}
\caption{Visualization results of our model in RealEstate10K~\cite{zhou2018re10k} dataset.}
\label{fig:re10k-supp}
\vspace{-0.5cm}
\end{figure*}

We provide additional visualizations from the RealEstate10K ~\cite{zhou2018re10k} dataset. As shown in Figure~\ref{fig:re10k-supp}, our model produces novel view synthesis results with fewer artifacts, particularly in areas that are challenging to match.

\subsection{More Results in ACID}
\begin{figure*}[htp]
\centering
\includegraphics[width=0.95\textwidth]{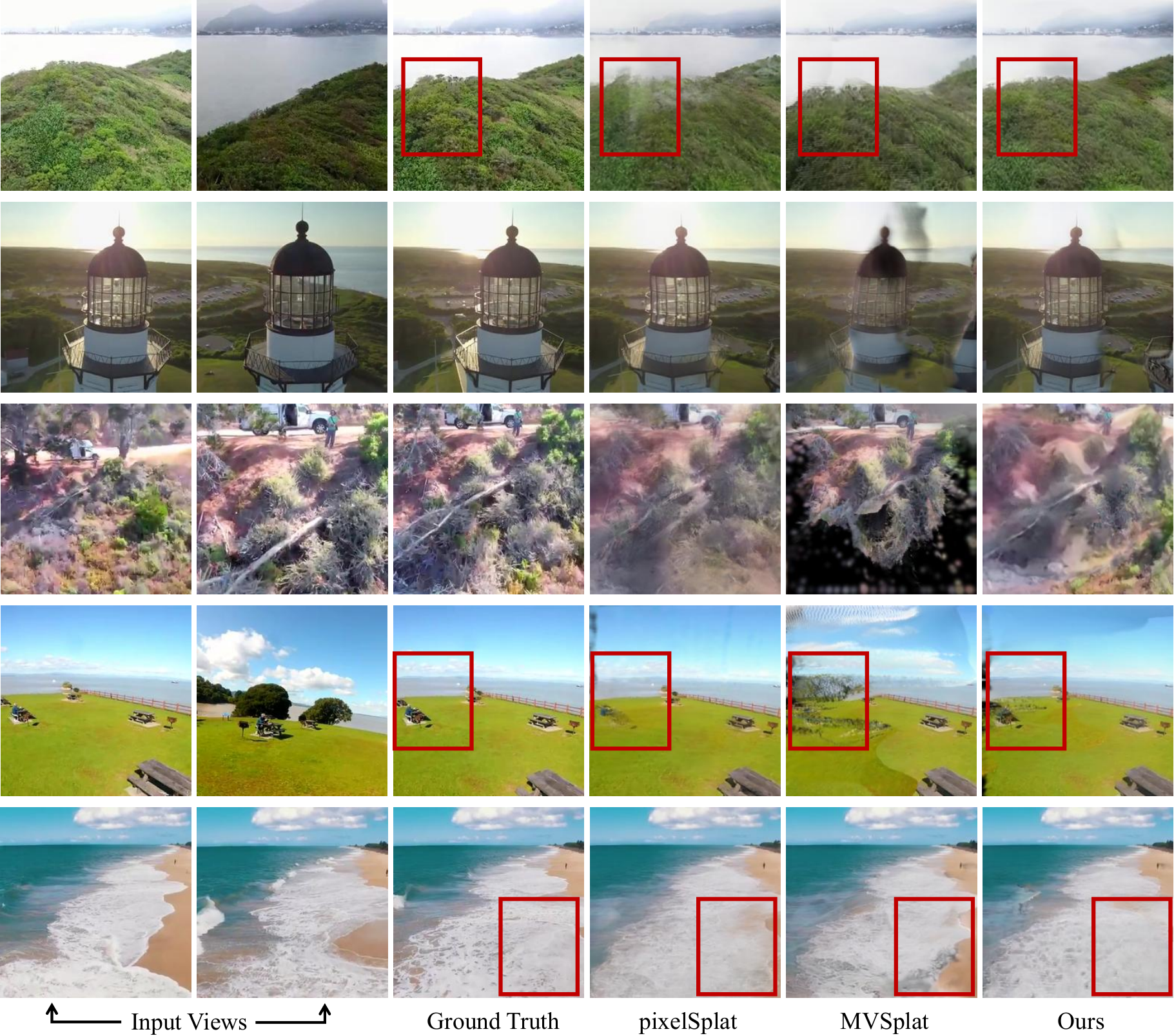}
\caption{Visualization results of our model in ACID~\cite{liu202acid} dataset.}
\label{fig:acid-supp}
\vspace{-0.5cm}
\end{figure*}

We provide additional visualizations from the ACID~\cite{liu202acid} dataset. As shown in Figure~\ref{fig:acid-supp}, our model gets the novel view synthesis results with fewer artifacts, especially in challenging real-world scenarios.

\subsection{More Results in DTU}
\begin{figure*}[htp]
\centering
\includegraphics[width=0.95\textwidth]{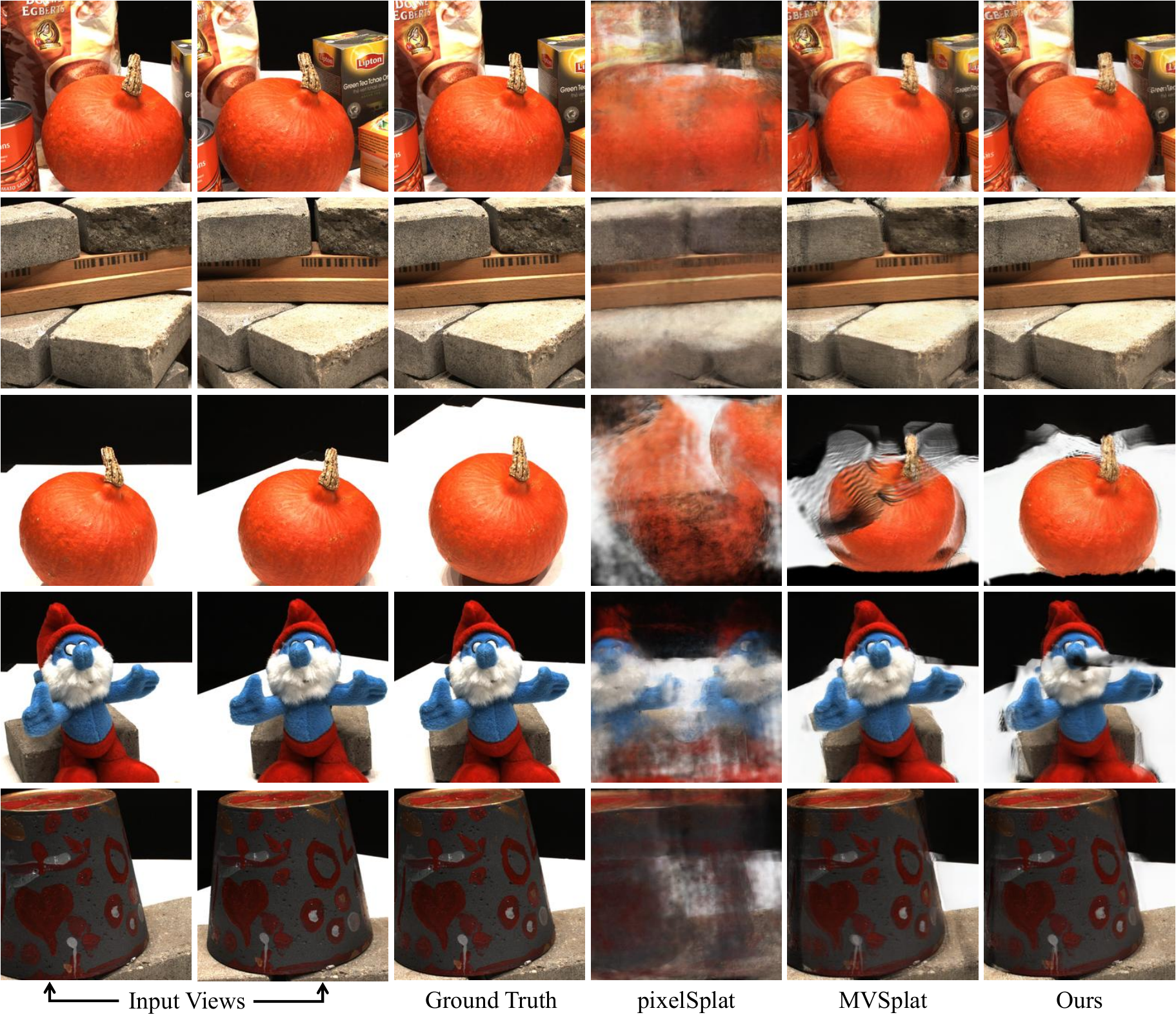}
\caption{Visualization results of our model in DTU~\cite{jensen2014dtu} dataset.}
\label{fig:dtu-supp}
\vspace{-0.5cm}
\end{figure*}

We perform zero-shot tests on the DTU dataset \cite{jensen2014dtu} using  models trained on the RealEstate10K dataset \cite{zhou2018re10k}.
Our method demonstrates the better performance in novel view synthesis, particularly in the low-texture regions, as shows in Figure~\ref{fig:dtu-supp}. Overall, TranSplat exhibits the strong cross-dataset generalization ability.


\end{document}